\documentclass{article}
\usepackage[T1]{fontenc}
\usepackage{float}   
\usepackage{spconf,amsmath,amssymb,graphicx,booktabs,multirow,array,url}
\usepackage[hidelinks]{hyperref}

\usepackage{xcolor}
\usepackage{stfloats}

\makeatletter
\def\fnum@figure{{\bfseries Figure~\thefigure}}
\makeatother
\newcommand{\bt}[1]{\textbf{#1}}
\newcommand{\pl}{\ensuremath{\tilde{y}}}

\title{Pseudo-Labels as Rewards, Not Targets: \\
       Group-Relative Policy Optimization for Semi-Supervised ASR}

\name{\begin{tabular}{@{}c@{}}Shashi Kumar$^{1,2}$, Yanis Labrak$^{1}$, Sergio Burdisso$^{1}$, Esaú Villatoro-Tello$^{1}$ \\ Petr Motlicek$^{1,3}$, Kadri Hacioglu$^{4}$, Andreas Stolcke$^{4}$\end{tabular}}

\address{$^{1}$Idiap Research Institute, Martigny, Switzerland;
         $^{2}$EPFL, Lausanne, Switzerland \\
         $^{3}$Brno University of Technology, Czech Republic;
         $^{4}$Uniphore, U.S.A.}

\begin{document}
\ninept
\maketitle
%
\begin{abstract}
Adapting ASR to a domain with only untranscribed audio typically relies on pseudo-labeling, a strong teacher transcribes the audio and a student is fine-tuned on it. Because supervised fine-tuning (SFT) treats every pseudo-label token as a target, the student inherits the teacher's errors. We show that using the same pseudo-labels as a reward reference for Group Relative Policy Optimization (GRPO) is more robust to label noise. In the banking domain, GRPO yields 29.3\% to 56.2\% relative WER improvements over SFT on Whisper-generated labels, and the advantage holds across four Whisper sizes, four teacher families, three domains and four languages. While SFT reproduces almost 100\% of teacher errors verbatim, GRPO passes through only 27\%. Rollout statistics explain why, erroneous pseudo-label words rarely appear in on-policy rollouts, so group-relative advantages stay aligned with the clean transcript, whereas SFT forces them into the policy's support and collapses the reward variance GRPO relies on.
\end{abstract}


\section{Introduction}
\label{sec:intro}
Automatic speech recognition (ASR) models degrade sharply outside the domain
they were trained on, yet the data available for adaptation rarely comes with
transcripts. In practice, a customer provides hours of in-domain audio and
nothing more. The standard remedy is self-training
\cite{manohar2018semisup,kahn2020selftraining,synnaeve2020e2e,lugosch2022multilingual}, in which a strong
off-the-shelf teacher such as Whisper \cite{radford2023whisper} transcribes
the audio and a student that fits the production constraints is trained on
the pseudo-labels (PL) with supervised fine-tuning (SFT) to adapt it
to the target acoustic conditions.

\vspace{1mm}

This recipe is simple and effective but structurally flawed, because
cross-entropy maximizes the likelihood of each pseudo-label token regardless
of whether it is correct. Every teacher error thus becomes a supervised
target, and deep networks readily memorize such targets
\cite{zhang2017rethinking,arpit2017memorization}, so the student tracks, and
often amplifies, the teacher's error profile. Existing work addresses this
problem by improving or filtering the labels while leaving the cross-entropy
loss unchanged. Iterative pseudo-labeling \cite{xu2020ipl}, slimIPL
\cite{likhomanenko2021slimipl}, and Kaizen \cite{manohar2021kaizen}, which
uses an exponential moving average of the student as teacher, regenerate
labels as the student improves, while noisy student training
\cite{park2020noisystudent} adds augmentation and scale. Labels can also be
improved before training by fusing several ASR systems and correcting their
output with a speech LLM \cite{prakash2025multiasr}, or pruned through data
selection based on domain classifiers, WER prediction, named-entity checks, or
multi-model consensus
\cite{rangappa2025domainclf,rangappa2025multistage,carofilis2025incremental}.
Replacing hard pseudo-labels with soft token distributions can make training
collapse \cite{pmlr-v187-likhomanenko23a}. All of these approaches still train
the student to reproduce the teacher's output.

\vspace{0.5mm}

An alternative is to change the objective and leave the label set untouched.
Sequence-level criteria such as minimum WER training
\cite{prabhavalkar2018mwer} and policy-gradient methods \cite{zhou2018policy}
have long been used for ASR, but always with gold transcripts. Group Relative
Policy Optimization (GRPO) \cite{shao2024deepseekmath}, developed for LLM reasoning \cite{deepseek2025r1}, simplifies PPO \cite{schulman2017ppo} by dropping its learned critic, the network that estimates the expected reward of each output.
Instead, several hypotheses are sampled for the same input and
each one is judged against the group's average reward. It has recently been applied to LLM-based ASR with rewards
computed against ground-truth references \cite{shivakumar2025grpoasr} or
derived from LLM feedback \cite{ling2025customizingspeechrecognitionmodel}, and to domain adaptation
from synthetic speech, where the reward still relies on the true transcript of
the synthetic audio \cite{kumar2026grpo}. In this work, every pseudo-label is kept but serves
as the reference for a WER-based GRPO reward rather than as a target for
maximum likelihood. GRPO has not yet been studied with such an unreliable
reference, even though it is the only relevant regime when no transcripts
exist.

\vspace{0.5mm}

The robustness of this approach comes from on-policy sampling rather than
from the reward function. Because the gradient acts only on
sequences the student generates, an erroneous pseudo-label word can be
reinforced only if the student already produces it.
This mechanism differs
from the one underlying most noise-robust training, in which networks fit
clean patterns before memorizing noise \cite{arpit2017memorization}. Here,
noise is filtered not by learning dynamics over time but by the support of the
sampling distribution at each step.

\vspace{2mm}
\noindent This paper reports four main findings:

\vspace{1mm}

\begin{itemize}\itemsep0pt
\item GRPO on pseudo-labels outperforms SFT at every level of label quality and is less sensitive to pseudo-label degradation (Section~\ref{sec:results}). On banking, GRPO trained on Whisper Tiny labels achieves 14.06\% WER, outperforming SFT trained on Large-V3 labels at 17.21\%. Thus, in this comparison, changing the training objective provides a larger benefit than improving the teacher.
\vspace{1mm}
\item This advantage holds across teacher families and conversational domains, and extends to low-resource languages once SFT has made RL applicable (Section~\ref{sec:results}).
\vspace{1mm}
\item Running SFT before GRPO is counterproductive with noisy labels, unless the student model cannot produce the target script (Section~\ref{sec:results}).
\vspace{1mm}
\item On-policy sampling explains these results. Label errors rarely enter GRPO's rollouts, whereas SFT places them in nearly every rollout and removes the reward variance GRPO needs (Section~\ref{sec:analysis}).
\end{itemize}

\section{Method}
\label{sec:method}

\subsection{Setup}
The available data consist of untranscribed target-domain audio
$\mathcal{D}=\{x_n\}_{n=1}^{N}$, and a teacher $T$ produces pseudo-labels $\pl_n = T(x_n)$. The student
policy $\pi_\theta$ is initialized from a student ASR model $\pi_{\theta_0}$
trained on out-of-domain data (LibriSpeech 960 hours \cite{panayotov2015librispeech}).
Gold transcripts are reserved for evaluation, diagnostic analysis, and separate gold-label training baselines.

\subsection{Supervised Fine-Tuning}
The conventional recipe minimizes the token-level cross-entropy of the
pseudo-label,
\begin{equation}
\mathcal{L}_{\text{SFT}}(\theta) = -\sum_{n}\sum_{t}
   \log \pi_\theta\!\left(\pl_{n,t}\mid \pl_{n,<t}, x_n\right).
\label{eq:sft}
\end{equation}
Every token of $\pl_n$ contributes to the loss whether or not it is correct,
and teacher forcing conditions later predictions on erroneous tokens as if
they were correct.

\subsection{Group Relative Policy Optimization}
For each utterance, a group of $G{=}4$ hypotheses $\{y^{(i)}\}_{i=1}^{G}$ is
sampled from $\pi_{\theta_{\text{old}}}(\cdot\mid x_n)$, and each one is scored
against the pseudo-label,
\begin{equation}
r_i = 1 - \mathrm{WER}\!\left(y^{(i)},\pl_n\right).
\label{eq:reward}
\end{equation}
The advantages are then normalized within the group,
\begin{equation}
\hat{A}_i = \frac{r_i - \mu_r}{\sigma_r + \epsilon},
\qquad
\mu_r = \tfrac{1}{G}\textstyle\sum_j r_j, \quad
\sigma_r = \mathrm{sd}\!\left(r_1,\dots,r_G\right),
\label{eq:adv}
\end{equation}
and $\theta$ is updated with the standard clipped surrogate objective of
\cite{shao2024deepseekmath} using $\hat{A}_i$, with clipping range
$\varepsilon$ and a KL penalty of weight $\beta$ that anchors $\pi_\theta$ to a
reference policy $\pi_{\text{ref}}$, taken as the policy at the start of the
RL stage.

\subsection{Why on-policy sampling should filter label noise?}
\label{sec:why} The two objectives distribute their gradient differently.
Equation~\ref{eq:sft} (SFT) acts on $\pl_n$ unconditionally, whereas GRPO acts only on sequences sampled from the policy itself. Suppose that $\pl_n$ contains an erroneous word $w$ in place of the correct word $w^\star$, and let $p$ be the probability that a rollout reproduces $w$.
Every rollout that outputs $w^\star$ is penalized at that position. When all rollouts in a group do so, they all receive the same penalty, which shifts every reward by the same amount and is cancelled by the subtraction of $\mu_r$ in Equation~\ref{eq:adv}.
The advantages therefore depend only on the positions where the rollouts differ from one another, not on those where they all agree, whether the pseudo-label is correct there or not. At a teacher error, the rollouts differ only if at least one of them reproduces $w$, which is rare when $p$ is small. Cross-entropy admits no such cancellation, since there the error is a target rather than an offset.

Rollouts that do contain $w$ are reinforced, so the error is not eliminated, but its gradient mass scales as $O(p)$ rather than $O(1)$. The mechanism is therefore self-limiting rather than self-correcting. 

It requires $p$ to be small at the outset, but not the student model to be accurate. A model with a high WER is often wrong, yet rarely wrong in exactly the same way as the teacher, because two different models tend to make different errors.

\section{Experimental Setup}
\label{sec:setup}

\noindent\textbf{Student model.} The student is the speech-LLM of
\cite{kumar2026grpo}, in which a frozen WavLM-Large encoder \cite{chen2022wavlm} feeds
a trainable projector, which consists of two linear layers with a ReLU, downsampling the
acoustic sequence by a factor of 5, following \cite{10.1609/aaai.v39i23.34666} approach, into a frozen Llama-3.2-1B-Instruct
\cite{grattafiori2024llama3} adapted with LoRA \cite{hu2022lora}
($r{=}16$, $\alpha{=}32$). Only the projector and the LoRA adapters are
trained. The student model is fine-tuned on LibriSpeech 960 hours
\cite{panayotov2015librispeech}.

\vspace{1mm}
\noindent\textbf{Target domains.} We focus on conversational domains where transcripts are scarce in practice. Customer calls in banking, healthcare, and insurance contain specialized vocabulary and sensitive personal information, so transcribing them requires expert annotators working under strict confidentiality, which makes large labeled sets costly to obtain. The primary target domain is conversational banking data from DefinedAI~\cite{kumar2026distillingconversationsabstractcompression,carofilis2026textonly,kumar2026grpo}, consisting of English customer--agent telephone calls, with approximately 54~h of training audio and 6.55~h of test audio. Healthcare and insurance conversations from the same provider serve as additional domains, with 11.51~h and 27.60~h of training audio (4.6K and 11K utterances) and 1.11~h and 3.24~h of test audio (488 and 1.3K utterances), respectively.

\vspace{1mm}
\noindent\textbf{Multilingual data.} Low-resource languages are a second setting where labels are scarce, since both existing transcribed corpora and qualified transcribers are limited. Common Voice 26~\cite{ardila2020commonvoice} provides data for Hindi, Hebrew, Vietnamese, and Serbian, with 5.86~h, 2.28~h, 2.33~h, and 1.97~h of training audio and 4.75~h, 1.34~h, 1.74~h, and 1.83~h of test audio, respectively. Their Whisper Large-V3 pseudo-labels have WERs of 26.60\%, 24.75\%, 6.08\%, and 17.29\%, respectively.

\vspace{1mm}
\noindent\textbf{Teachers.} Whisper (Tiny, Small, Medium and Large-V3) \cite{radford2023whisper} provide a controlled quality ladder within a single architecture family. Scored against the gold transcripts of the banking training audio, their pseudo-labels have WERs of 26.42, 17.23, 16.61, and 14.61, respectively, an 11.81-point spread exploited throughout Section~\ref{sec:results}. Qwen3 ASR 1.7B \cite{qwen3asr2026}, Phi-4 Multimodal IT \cite{abouelenin2025phi4mm} and Audio Flamingo~3 \cite{goel2025audioflamingo3} serve as additional teachers to test whether the effect stems from the method or from a teacher family and very high quality labels.

\vspace{1mm}
\noindent\textbf{Training.} The GRPO configuration follows \cite{kumar2026grpo}, with group size $G{=}4$, sampling temperature $\tau{=}0.8$, at most 128 new tokens per training rollout, KL coefficient $\beta{=}0.04$, and clipping range $\varepsilon{=}0.2$. Both SFT and GRPO use AdamW with BF16 precision, a batch size of 16, 100 warmup steps, gradient-norm clipping at 1.0, and 5 epochs, with learning rates of $1{\times}10^{-4}$ for SFT and $2{\times}10^{-5}$ for GRPO. All training runs on a single NVIDIA H100 80\,GB GPU. Test-time decoding is greedy (beam size 1) with a 256-token limit.

\vspace{-2mm}
\section{Results}
\label{sec:results}
\vspace{-1mm}

\noindent\textbf{GRPO absorbs label degradation while SFT inherits it.} Table~\ref{tab:main} reports the core result. From Large-V3 to Tiny, label WER rises by 11.81 points, but GRPO student WER rises by only 1.89 (12.17 to 14.06), whereas SFT student WER rises by 14.90 (17.21 to 32.11).
Across the four teachers, each additional point of label WER increases student WER by 1.31 points on average with SFT, 0.56 with SFT$\rightarrow$GRPO, and only 0.16 with GRPO.
GRPO thus outperforms SFT with every teacher, and its margin grows from 5.04 points with Large-V3 labels to 18.05 with Tiny labels. GRPO trained on the weakest labels (Tiny) even beats SFT trained on the strongest (Large-V3) by 3.15 points (14.06 vs.\ 17.21). The choice of teacher therefore matters much less under GRPO, which is useful when no transcripts are available to compare teachers.

\begin{table}[H]
\centering\ninept\setlength{\tabcolsep}{3.4pt}
\caption{Performance on the banking test set of DefinedAI. Students are trained on pseudo-labels from four Whisper teachers, from weakest to strongest, or on gold transcripts as a topline.}
\vspace{3mm}
\label{tab:main}
\begin{tabular}{llccl}
\toprule
PL Source & Method & WER & CER & INS/DEL/SUB \\
\midrule
\textit{LS960 (student)} & SFT & 66.44 & 46.73 & 42.18/5.68/18.59 \\
\midrule
\multirow{3}{*}{\textit{Tiny}}
 & SFT & 32.11 & 23.71 & 15.17/4.91/12.03 \\
 & GRPO & \bt{14.06} & \bt{10.68} & \bt{1.87/5.69/6.49} \\
 & SFT$\rightarrow$GRPO & 22.77 & 16.71 & 6.20/5.77/10.80 \\
\midrule
\multirow{3}{*}{\textit{Small}}
 & SFT & 20.23 & 15.64 & 7.95/4.00/8.28 \\
 & GRPO & \bt{12.71} & \bt{9.64} & \bt{1.95/4.84/5.91} \\
 & SFT$\rightarrow$GRPO & 17.39 & 13.43 & 4.90/4.60/7.89 \\
\midrule
\multirow{3}{*}{\textit{Medium}}
 & SFT & 18.11 & 14.06 & 6.30/4.47/7.33 \\
 & GRPO & \bt{12.58} & \bt{9.54} & \bt{1.98/4.67/5.93} \\
 & SFT$\rightarrow$GRPO & 17.17 & 13.44 & 5.03/4.83/7.30 \\
\midrule
\multirow{3}{*}{\textit{Large-V3}}
 & SFT & 17.21 & 13.60 & 6.25/3.77/7.19 \\

 & GRPO & \bt{12.17} & \bt{9.29} & \bt{2.12/4.32/5.73} \\
 & SFT$\rightarrow$GRPO & 16.31 & 12.91 & 5.35/3.95/7.01 \\
\midrule
\midrule
\multirow{3}{*}{\textit{Gold}}
 & SFT & 10.27 & 7.11 & 2.99/2.13/5.14 \\

 & GRPO & 10.78 & 7.68 & 1.79/3.50/5.49 \\
 & SFT$\rightarrow$GRPO & \bt{9.49} & \bt{6.66} & \bt{1.98/2.50/5.01} \\
\bottomrule
\end{tabular}
\end{table}


\vspace{1mm}
\noindent\textbf{SFT before GRPO is harmful.} SFT$\rightarrow$GRPO closes only part of the gap between SFT and direct GRPO (22.77 vs.\ 14.06 with Tiny labels), which inverts the usual LLM post-training recipe. With gold labels, the usual ordering does hold (SFT$\rightarrow$GRPO gives the best WER, 9.49), which isolates label noise as the cause. SFT is a good initialization when targets are correct and liability when they are not.

\vspace{1mm}
\noindent\textbf{GRPO suppresses insertions.} The error breakdown reveals a
consistent signature. GRPO sharply reduces insertions (15.17 $\rightarrow$
1.87 with Tiny labels) at the cost of slightly more deletions (4.91
$\rightarrow$ 5.69). A WER-based reward directly penalizes hallucinated
content, so the policy learns to be conservative, consistent with previous work \cite{shivakumar2025grpoasr}.

\begin{table}[H]
\centering\ninept
\caption{Performance (\%WER) on the banking test set of DefinedAI with pseudo-labels from three additional teacher families.}
\vspace{3mm}
\label{tab:teachers}
\begin{tabular}{lccc}
\toprule
Teacher & SFT & \bt{GRPO} & SFT$\rightarrow$GRPO \\
\midrule
\textit{Qwen3 ASR 1.7B} & 13.83 & \bt{11.35} & 13.41 \\
\textit{Phi-4 Multimodal IT} & 14.34 & \bt{11.81} & 14.76 \\
\textit{Audio Flamingo 3} & 14.96 & \bt{11.83} & 14.32 \\
\bottomrule
\end{tabular}
\end{table}

\noindent\textbf{The effect is teacher-agnostic.} Table~\ref{tab:teachers} repeats the comparison with three additional teachers spanning different architectures. Qwen3 ASR, Phi-4 and Audio Flamingo~3 produce pseudo-labels with WERs of 11.05, 12.46, and 12.44, respectively, all lower than the 14.61 of Whisper Large-V3. GRPO outperforms SFT in all four cases. With Qwen3 ASR pseudo-labels, GRPO achieves 11.35\% test WER compared with 13.83\% for SFT, a reduction of 2.48 percentage points.

\vspace{1mm}
\noindent\textbf{The effect transfers across domains.} Table~\ref{tab:domains} shows the same pattern on healthcare (18.72 $\rightarrow$ 14.64 WER, against a gold-label reference of 12.28) and insurance (19.59 $\rightarrow$ 15.70, against 14.02). With gold labels, GRPO is marginally worse than SFT in WER in both domains, indicating that the advantage is specific to noisy references rather than a general property of the objective. Gold-label runs also quantify the performance penalty from replacing transcripts with pseudo-labels. On banking, this penalty is 6.94 WER points for SFT but only 1.39 for GRPO with Large-V3 labels. The same pattern holds in healthcare (6.44 versus 2.04 points) and insurance (5.57 versus 1.35 points). Thus, across all three domains, GRPO stays closer to its corresponding gold-label result, reinforcing that its main benefit is robustness to imperfect supervision.

\begin{table}[H]
\centering\ninept
\caption{Performance on the DefinedAI healthcare and insurance test sets. Pseudo-labels come from Whisper Large-V3, and gold-label results serve as a topline.}
\vspace{3mm}
\label{tab:domains}
\begin{tabular}{ll cc cc}
\toprule
 & & \multicolumn{2}{c}{Gold labels} & \multicolumn{2}{c}{Pseudo-labels} \\
\cmidrule(lr){3-4}\cmidrule(lr){5-6}
Domain & Method & WER & CER & WER & CER \\
\midrule
\multirow{2}{*}{\textit{Healthcare}}
 & SFT       & \textbf{12.28} & 8.40  & 18.72 & 14.32 \\
 & GRPO & 12.60 & \textbf{8.39}  & \bt{14.64} & \bt{10.91} \\
\midrule
\multirow{2}{*}{\textit{Insurance}}
 & SFT       & \textbf{14.02} & 10.27 & 19.59 & 15.53 \\
 & GRPO & 14.35 & \textbf{10.24} & \bt{15.70} & \bt{12.12} \\
\bottomrule
\end{tabular}
\end{table}


\noindent\textbf{Multilingual.}
Table~\ref{tab:multiling} extends the study to four Common Voice languages with the same student configuration. With pseudo-labels, adding GRPO after SFT reduces CER for every language (e.g., 58.78 $\rightarrow$ 49.92 for Vietnamese), and the resulting systems match or outperform their gold-label counterparts. With gold labels, GRPO helps for all languages except Hindi. Absolute CERs are high throughout, as expected at this scale, since each language has under 6\,h of training audio, and comparable LLM-based ASR systems report CERs in the same range \cite{zhang-etal-2026-language}. The relevant comparison is therefore between methods within a language, not the absolute level.
\vspace{-1mm}
\begin{table}[H]
\centering\ninept
\caption{CER (\%) on the Common Voice 26 test sets for four low-resource
languages, each with under 6\,h of training audio. Pseudo-labels come from
Whisper Large-V3. CER is used instead of WER because these scripts are
non-Latin or heavily diacritized. Direct GRPO is omitted because the
English-trained student model cannot produce the target scripts
(Section~\ref{sec:results}).}
\vspace{3mm}
\label{tab:multiling}
\begin{tabular}{lcccc}
\toprule
 & \multicolumn{2}{c}{Gold labels} & \multicolumn{2}{c}{Pseudo-labels} \\
\cmidrule(lr){2-3}\cmidrule(lr){4-5}
Language & SFT & SFT$\rightarrow$GRPO & SFT & SFT$\rightarrow$GRPO \\
\midrule
\textit{Hindi}      & \textbf{27.01} & 28.26 & 24.27 & \textbf{22.94} \\
\textit{Hebrew}     & 46.40 & \textbf{37.33} & 40.03 & \textbf{36.89} \\
\textit{Vietnamese}  & 57.64 & \textbf{51.02} & 58.78 & \textbf{49.92} \\
\textit{Serbian}    & 34.31 & \textbf{30.68} & 35.01 & \textbf{30.17} \\
\bottomrule
\end{tabular}
\end{table}

\noindent\textbf{Where the mechanism breaks down.} Direct GRPO is omitted for
these languages because it is not applicable. The student model is trained on
English, and its rollouts are English-like phonetic renderings in Latin script
rather than text in the target script. Every hypothesis therefore scores above
100\% WER against the pseudo-label, and the group-relative advantage no longer
separates better from worse transcriptions; it merely favors whichever
rollout most resembles the target language. This is the support argument of
Section~\ref{sec:why} in reverse. GRPO filters label noise because the student
policy already places probability mass on the correct word, so when the
correct output lies entirely outside that support, on-policy sampling has
nothing to select. A short SFT stage establishes the target script and makes
the subsequent RL stage well-posed. This is the only regime observed in
which SFT first is necessary rather than harmful.

\section{Analysis}
\label{sec:analysis}

The results raise two questions. Why does GRPO outperform SFT on the same
pseudo-labels, and why does SFT$\rightarrow$GRPO remain far behind direct
GRPO? Section~\ref{sec:why} attributes both observations to on-policy
sampling. GRPO can reinforce a label error only when the current policy
samples it, whereas SFT trains on every token unconditionally. This
explanation is tested by tracking individual label errors in the final outputs
and in the GRPO rollouts. Gold transcripts serve only to identify pseudo-label
errors in this post-hoc analysis, and utterances are binned by their number of
errors.

\begin{figure}[H]
\centering
\includegraphics[width=\linewidth]{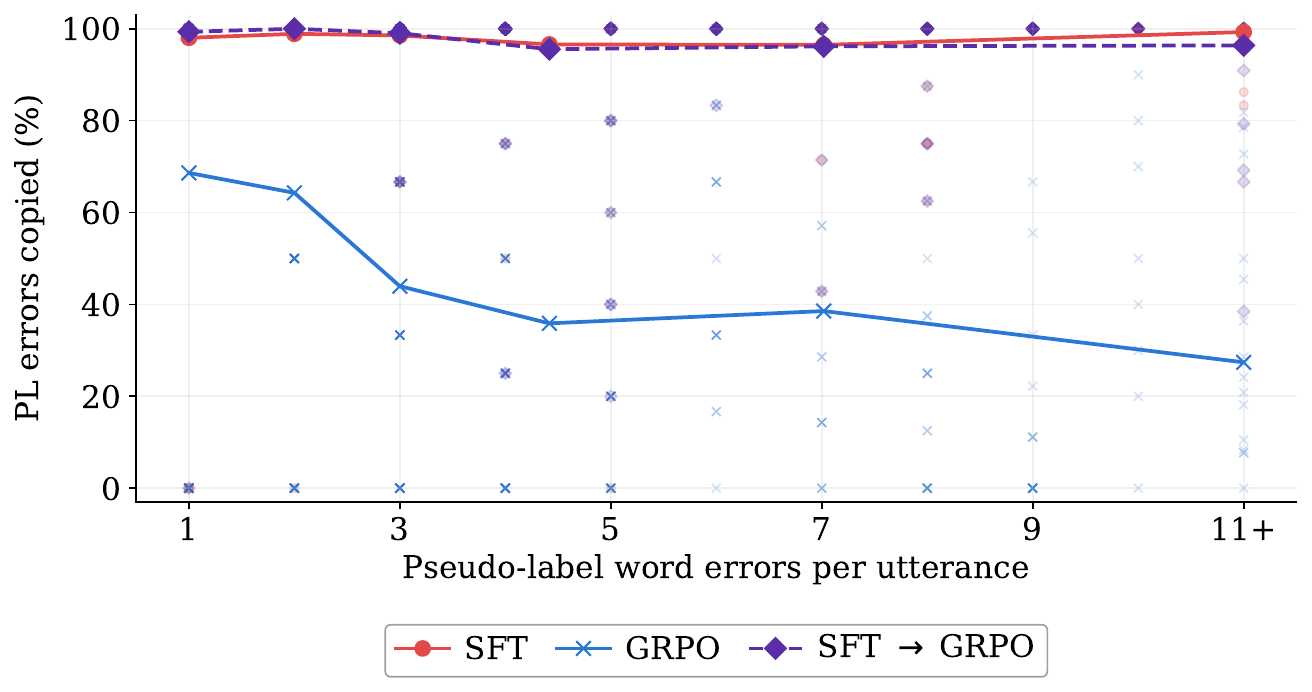}
\caption{Fraction of erroneous pseudo-label (PL) words that the student reproduces verbatim in its final output, as a function of the number of word errors in the utterance's pseudo-label.}
\label{fig:copyrate}
\end{figure}

\begin{figure*}[b]
\centering
\includegraphics[width=0.49\textwidth]{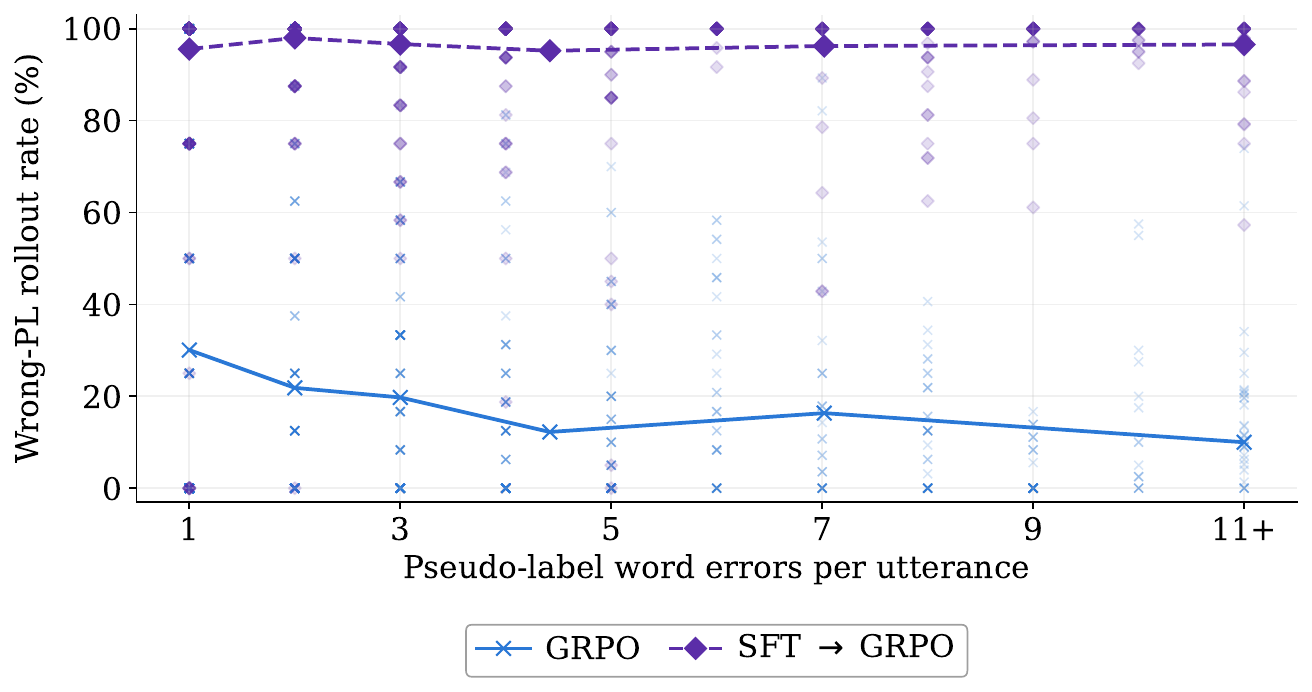}\hfill
\includegraphics[width=0.49\textwidth]{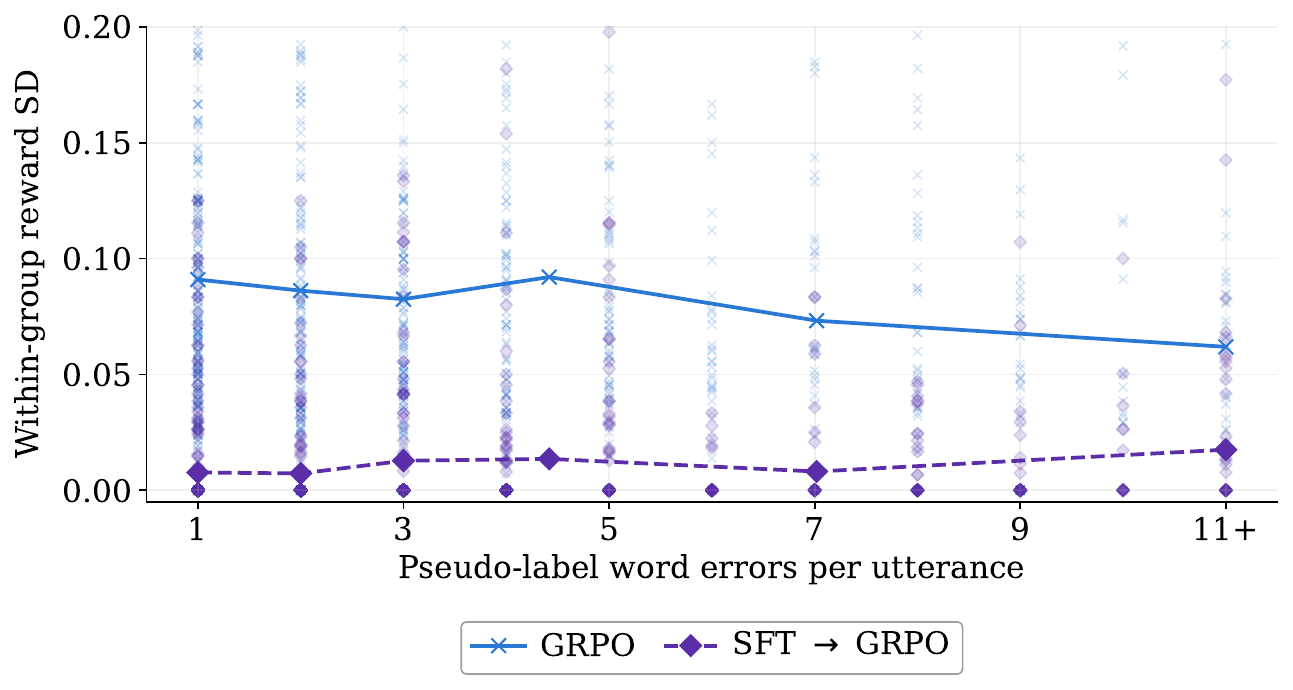}
\caption{Rollout statistics as a function of the number of PL word errors per
utterance. \bt{Left:} fraction of the $G{=}4$ rollouts per utterance that
contain the erroneous PL word. \bt{Right:} within-group standard deviation of
the reward across the $G{=}4$ rollouts.}
\label{fig:rollstats}
\end{figure*}

\vspace{-1mm}
\noindent\textbf{GRPO copies fewer teacher errors as label quality worsens.}
Figure~\ref{fig:copyrate} measures how often an erroneous pseudo-label word survives into the student's final output. SFT copies nearly 100\% of these errors at every corruption level, and SFT$\rightarrow$GRPO closely follows it. Running SFT first thus eliminates GRPO's advantage before GRPO can exploit it. Direct GRPO, in contrast, copies progressively fewer errors as pseudo-label quality degrades, from 68\% in the one-error bin to 27\% in the 11+ error bin. This reflects the on-policy mechanism of Section~\ref{sec:why} compounding across positions. To reproduce $k$ independent pseudo-label errors, a rollout must miss the correct word $k$ times within the same sequence, so heavily corrupted utterances are filtered most strongly.

\vspace{1mm}
\noindent\textbf{Label errors rarely enter GRPO's rollouts.}
Figure~\ref{fig:rollstats} (left) identifies where the effect in Figure~\ref{fig:copyrate} originates. Before direct GRPO, an erroneous pseudo-label word appears in only 10 to 30\% of the rollouts; after SFT, it appears in more than 95\%. An erroneous word absent from the rollouts penalizes all of them equally, so its penalty cancels under group-relative normalization and the update is driven by the mostly correct parts of the pseudo-label. SFT, in contrast, trains directly on every erroneous word and makes it common in the rollouts of the subsequent GRPO stage. The student model’s high banking WER (66.44\%) therefore does not preclude useful GRPO updates. What matters is how often its sampled hypotheses reproduce the particular errors in the pseudo-labels. The low observed overlap suggests that overall recognition accuracy alone does not determine susceptibility to copying teacher errors.


\vspace{1mm}
\noindent\textbf{SFT removes the reward variance needed to correct its errors.}
Direct GRPO maintains a within-group reward standard deviation of 0.06 to
0.09, whereas after SFT it drops to about 0.01 in Figure~\ref{fig:rollstats} (right).
Because SFT-initialized rollouts reproduce nearly the same errors, they
receive nearly identical rewards and provide little signal to correct them.
The residual variance lets SFT$\rightarrow$GRPO improve over SFT, but not
enough to match direct GRPO.


\vspace{1mm}
\noindent Together, these diagnostics answer both questions. GRPO outperforms SFT because on-policy sampling rarely reproduces label errors, so noise receives little gradient. SFT$\rightarrow$GRPO improves over SFT for the same reason, but only weakly, because little reward variance survives SFT. It never reaches direct GRPO because SFT has already placed the errors within the policy's support and removed most of the signal needed to correct them.

\section{Conclusion}
\label{sec:conclusion}

Using pseudo-labels as reward references for GRPO enables robust ASR adaptation without target-domain transcripts, reducing WER by up to 18.1 points relative to SFT. Our analysis attributes this robustness to on-policy sampling, which limits reinforcement of pseudo-label errors. These findings establish the importance of initialization: direct GRPO is effective when the student model supports the target language, whereas SFT is essential when the output format of the model is not  compatible with the output format of the  task. 
Future work will combine GRPO with iterative pseudo-labeling~\cite{xu2020ipl}, reusing the student as teacher, and restrict initial SFT to high-confidence utterances to avoid copying teacher errors.




\newpage

\section{Acknowledgments}
This work was supported by Idiap Research Institute and Uniphore collaboration project. Part of this work was also supported by EU Horizon 2020 project ELOQUENCE (grant number 101070558).

\section{Compliance with Ethical Standards}
This research was conducted retrospectively using human subject speech data made available in open access or under institutional license (CommonVoice 26 and DefinedAI). Ethical approval was not required, as confirmed by the licenses of these corpora.



\bibliographystyle{IEEEbib}
\bibliography{refs}

\end{document}